\def\year{2020}\relax
\documentclass[letterpaper]{article} 
\usepackage{aaai20}  
\usepackage{times}  
\usepackage{helvet} 
\usepackage{courier}  
\usepackage[hyphens]{url}  
\usepackage{graphicx} 
\usepackage{graphicx}  
\author{Yufan Luo, Li Xiao \thanks{Corresponding Author} \\
Institute of Computing Technology, Chinese Academy of Science\\
xiaoli@ict.ac.cn
}

\title{G-RCN: Optimizing the Gap between Classification and Localization Tasks for Object Detection }

\begin{document}

\maketitle
\begin{abstract}
Multi-task learning is widely used in computer vision. Currently, object detection models utilize shared feature map to complete classification and localization tasks simultaneously. By comparing the performance between the original Faster R-CNN and that with partially separated feature maps, we show that: (1) Sharing high-level features for the classification and localization tasks is sub-optimal; (2) Large stride is beneficial for classification but harmful for localization; (3) Global context information could improve the performance of classification. Based on these findings, we proposed a paradigm called Gap-optimized region based convolutional network (G-RCN), which aims to separating these two tasks and optimizing the gap between them. The paradigm was firstly applied to correct the current ResNet protocol by simply reducing the stride and moving the Conv5 block from the head to the feature extraction network,which brings  3.6 improvement of $AP_{70}$ on the PASCAL VOC dataset and 1.5 improvement of $AP$ on the COCO dataset for ResNet50.  Next, the new method is applied on the Faster R-CNN with backbone of VGG16,ResNet50 and ResNet101, which brings above 2.0 improvement of $AP_{70}$ on the PASCAL VOC dataset and above 1.9 improvement of $AP$ on the COCO dataset. Noticeably, the implementation of G-RCN only involves a few structural modifications, with no extra module added.
\end{abstract}

\section{Introduction}
The development of deep learning substantially improves the ability of computer to perform the image analysis, especially when recognizing and detecting objects. In recent years, the study of convolutional neural networks\cite{krizhevsky2012imagenet,simonyan2014very,szegedy2015going,he2016deep,girshick2015fast,ren2015faster} promotes the acquisition of super-human level recognition ability\cite{hu2018squeeze} on ImageNet\cite{deng2009imagenet} dataset and obtains 0.62 at $AP_{50}$ \cite{peng2018megdet} on COCO\cite{lin2014microsoft} detection dataset respectively. Meanwhile, the correlation between recognition and detection tasks has been caught to transfer feature extraction networks from ImageNet recognition task to various detection tasks\cite{girshick2014rich}. And also very deep feature extraction networks such as ResNet\cite{he2016deep} have been proven to bring performance boosting in both recognition and detection tasks simultaneously, which indicates the similarity between recognition and detection tasks.

However, although deep neural networks have achieved the super-human level performance in 1000 classes of object recognition task, the detection networks have not obtained comparable performance to human in 80 classes of object detection tasks, which gives the evidence that there exists gap between recognition and detection tasks. Object detection requires powerful classification performance together with capacity of precise localization to determine one in infinite candidate positions. Intuitively, classification task needs more abstract and comprehensive information to tell whether the object has some specific features, for example, the network may need to judge whether an object has hands when recognizing a human. On the contrary, localization task prefers detailed, location sensitive feature information, for example, we need to extract the specific location information of a finger in order to accurately localize a human. However, the widely used detection networks, including both one-stage\cite{redmon2016you,liu2016ssd} and two-stage models\cite{girshick2014rich,girshick2015fast,ren2015faster}, employ the paradigm with shared feature extraction neural networks which extract features that contribute to both the classification and localization tasks simultaneously. As shown later, this will be proven as a sub-optimal strategy, because of the huge gap between the classification and localization tasks.

Several researches have been proposed to optimize this gap. For example, \cite{dai2016r} claims that localization and classification require translational variant features and translational invariant features respectively, and proposes to generate position-sensitive score map to solve this distinction. \cite{cheng2018revisiting} constructs different classification head and localization head to perform localization and classification tasks separately. \cite{singh2018analysis} attributes the performance drop of the detection tasks to the large scale variation between objects, and then presents a novel training method directed against the scale mismatching during task transferring. But none of them discuss where the gap exactly is and propose a pertinent solution.

This work aims at exploring and optimizing the gap between the classification and localization tasks. We firstly separated the last block of the convolutional layers for the classification and localization tasks to demonstrate that sharing the high-level features for the classification and localization tasks is sub-optimal. By introducing the attention mechanism to obtain a more abstract feature map as well as reducing the stride of the convolutional layers respectively, we further demonstrated that: (1) Large stride is beneficial for classification but harmful for localization;  (2) Global context information could improve the performance of classification. Based on the observations, we proposed a novel paradigm, which we refer as  Gap-optimized region based convolutional network(G-RCN) to separate the classification and localization tasks and optimize the gap between them. Noticeably, only the last few convolutional layers need to be separated therefore the separation brings very little parameter growth.

The G-RCN was applied on the Faster R-CNN with backbone of VGG16, ResNet50 and ResNet101, and  tested on the PASCAL VOC and COCO dataset. The implementation of G-RCN brings significant performance improvement for all the backbones, with above 2.0 improvement of $AP_{70}$ on the PASCAL VOC dataset and above 1.9 improvement of $AP$ on the COCO dataset. We further explored that current protocol which places the Conv5 block of the ResNet as the head to improve performance is misleading, and correct it by reducing the stride and moving the Conv5 block back to the feature extraction network. Noticeably, the simple modification brings 3.6 improvement of $AP_{70}$ on the PASCAL VOC dataset and 1.5 improvement of $AP$ on the COCO dataset for ResNet50. Furthermore, since the conflict between classification and localization tasks is universal, we expect our paradigm to be applied to all the other state-of-the-art object detection models.

\section{Related Work}
R-CNN based detection models\cite{girshick2014rich,girshick2015fast,ren2015faster} bring standard paradigm of two stage detection networks which is widely used in practice.
Recent improvements on object detection can mainly be summarized as three directions. The first direction was to obtain more powerful representation by adding extra structures to extract additional information or enhanced features. Among them, methods of adding context information\cite{bell2016inside,he2017mask,gidaris2015object} are in the primary direction. For example, \cite{gidaris2015object} obtains more discriminative and diverse features by adding appearance features of different regions around targets and enhancing the representation ability by multi-task learning with segmentation task. Besides, the Mask R-CNN\cite{he2017mask} for the object detection task also uses the auxiliary information from segmentation task to achieve performance improvement. The popularity of the attention model\cite{vaswani2017attention,xu2015show} not only produces the Relation Network\cite{hu2018relation} which is the current best use of the context information on the object detection network, but also brings stronger feature representation and leads to performance boosting on detection, segmentation and classification tasks\cite{hu2018squeeze}.

Another way to improve network performance is by making better use of the training data. Hard example mining\cite{shrivastava2016training} is representative of this approach by learning from difficult samples. Also, some studies try to control the learning procedure by designing appropriate loss function of the network. For example, focal loss\cite{lin2017focal} prevents the learning bias of the network by designing a new cross-entropy function directed against imbalance between positive/negative samples and simple/hard samples.

The third way relies on improving the expressive capacity of the feature extraction network. For example, the DCN\cite{dai2017deformable} enables the network to extract more focused features by using deformable convolutional layers and deformable ROI pooling layers. The FPN\cite{lin2017feature} develops a top-down architecture and builds literal connections that allow each layer of the feature extraction network to have precise detailed information and rich semantic information simultaneously. SE-Net\cite{hu2018squeeze} utilizes the attention mechanism by assigning different importance weights to different channels. ResNet\cite{he2016deep},Inception-ResNet\cite{szegedy2017inception} and ResNetXt\cite{xie2017aggregated} construct deeper thus more powerful feature extraction networks and improve performance in both classification and detection tasks.

Our work is also related to DetNet\cite{li2018detnet}, which designs a backbone that is specifically for detection problems. Besides, \cite{cheng2018revisiting} suggests to use unshared parameter of heads while we suggest to use unshared feature extraction network.

\section{Exploring the Gap between Classification and Localization}
In this section, we demonstrated that the widely used detection paradigm which performs classification and localization tasks simultaneously is suboptimal. To that end, we conducted several experiments based on the Faster R-CNN \cite{ren2015faster} with backbone of VGG16\cite{simonyan2014very} as the baseline and several experiments was done to explore the gap. The super-parameters and settings were set the same as \cite{ren2015faster} in all the experiments unless announced specially. The modified architectures were all trained and validated based on the COCO 2014 detection datasets. Based on the findings, we modified the paradigm slightly and provide significant performance improvements without bells and whistles. As the development is based on the general region-based object detection paradigm, it can be extend to apply to other state-of-the-art object detection models.

\subsection{Separation Improves Performance}
\begin{figure*}
\begin{center}
\includegraphics[width=0.9\textwidth]{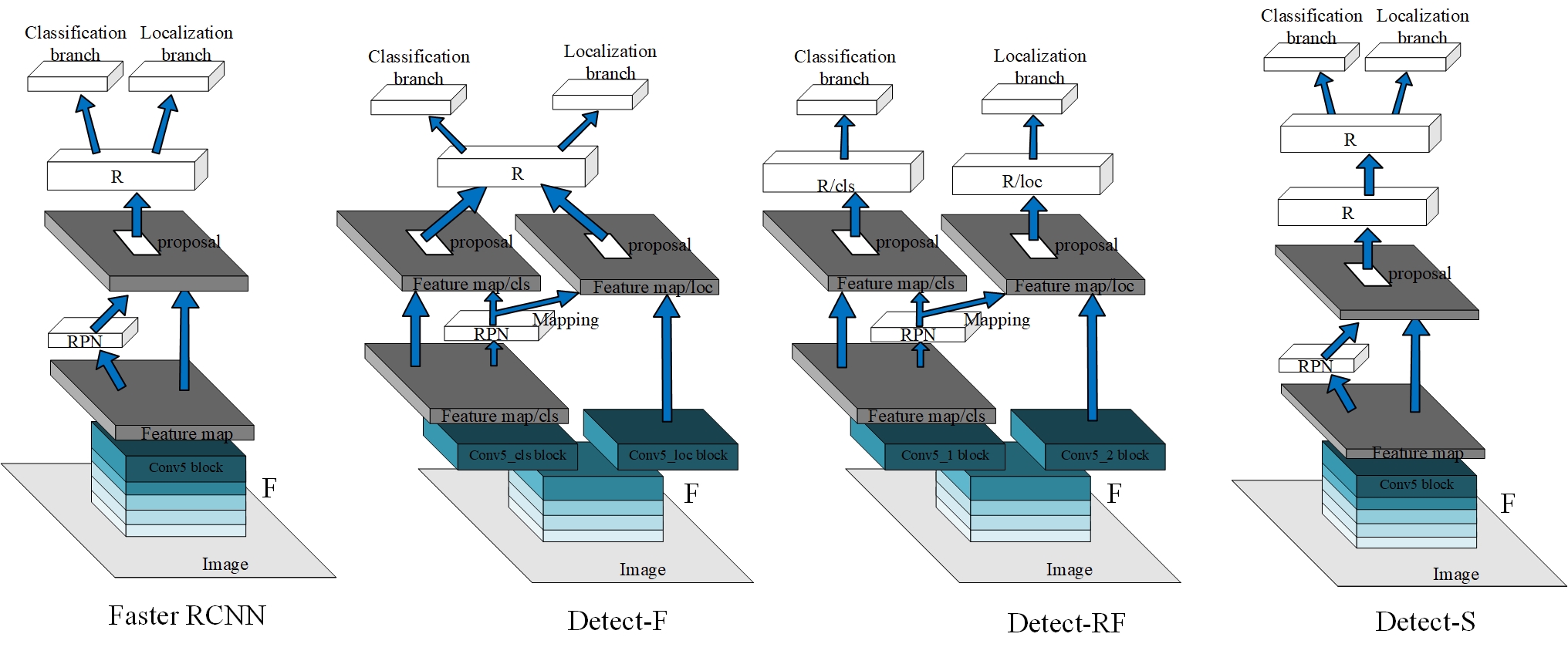}
\end{center}
   \caption{Models used to explore the gap between classification and localization.  F represents backbone and R represents head.}
\label{fig4}
\end{figure*}

We firstly conducted several experiments to prove that there are clear gaps between classification and localization tasks. We separated classification and localization tasks in different parts of detection models to observe the performance of each corresponding modified architecture.

\begin{table}
\begin{center}
\begin{tabular}{c|c c c }
 \hline
 & AP& $AP_{50}$& $AP_{75}$\\
 \hline
 Faster R-CNN & 21.3 &42.0 &19.8  \\
 detect-F & 21.8& 41.9& 20.7\\
 detect-RF& 21.9& 41.9& 20.9\\
 detect-S& 21.3& 41.6& 19.9\\
 \hline
\end{tabular}
\end{center}
\caption{Model performances to demonstrate the existence of gap between classification and localization.VGG16 is adopted as backbone.The structures of different models are shown in Figure \ref{fig4}.}
\label{tab1}
\end{table}

Three different models are developed to demonstrate the improvement of performance when separating the classification and localization tasks. All filters are initialized by the pre-trained network on ImageNet, including those filters that are separated for the two tasks.  As shown in Figure \ref{fig4}, the detect-F is constructed by using two different conv5 blocks for classification and localization tasks, and generates anchor proposals for the two tasks separately. The detect-RF adopts two independent multi-layer perceptions in the head for classification and localization respectively. The detect-S is constructed as a comparison with detect-RF, which adds another head but not separate the two tasks, and therefore it has the same level of parameters as the detect-RF.

It is worth mention that in detect-F and detect-RF, the original RPN  is firstly applied on the feature maps of the classification branches to generate anchors and predict a score for each anchor, and then the top 2000 anchors is selected and passed to the head R to perform classification. At the same time, the selected anchors are mapped onto the feature maps of the localization branches to perform localization.

Results are summarized in Table \ref{tab1}. Different from other related researches\cite{cheng2018revisiting}, we not only demonstrated the existence of gap between classification and localization, but also experimentally prove that the separation of these two tasks can be applied on the feature map level to improve performance. Interestingly, the detect-F has the similar performance as detect-RF, but only has a few convolutional layers more than the Faster R-CNN. On the contrary, the detect-S does not bring any performance improvement. This eliminates the argument that the performance boosting of the detect-RF is brought by the dramatic increasing of parameters.  It is worth mention that the separation of other convolutional blocks does not bring additional improvement, which suggests that classification and localization are both beneficial for low-level and middle-level feature extraction while brings distinction on the high-level features.

In the rest of this article, we will discuss what each task needs and where is the gap, and then proceed to further modify the detection paradigm aiming at each task.

\subsection{Global context on detection}
The influence of context information on detection remains a mystery.  We suggest that rich global context information, which adds more auxiliary and diversified information, can improve the classification significantly. On the contrary, as with intuition, rich global context information may confuse the localization process and bring no improvement. In the below experiments, we extract global context information based on \cite{vaswani2017attention} and enhance object features by combining the context information of the whole image.

The backbone generates feature map $G$ by filtering the whole image, and then the ROI pooling layer crops activations lay inside proposals and generate activation map $P$ with a fixed size. In our experiments, the VGG16 backbone is used and thus the dimensions of the feature maps are 512. As some global context information may not be relevant or useful , we adopted an attention procedure to select information that is beneficial for the proposal $P$ from the global context information $G$. In our experiments, $P$ which is the result of 7*7 ROI pooling of proposal constitutes a collection of 49 queries, each query is represented by a 512-dimensional vector. Then we perform $G$ with a 14*14 ROI pooling to obtain $V$ which constitutes a set of 196 key-value pairs (key and value are the same things here), each key-value pair is also a  512-dimension vector.   We used an attention model similar to \cite{hu2018relation}, which came from the multi-head dot-product attention module\cite{vaswani2017attention} to filter out noise of the global context information.

\begin{equation}
P = P + softmax(\frac{W_{P}PV^{T}W_{K}^{T}}{\sqrt{k}})W_{V}V
\end{equation}

\begin{table}
\begin{center}
\begin{tabular}{c|c c c }
 \hline
 & AP& $AP_{50}$& $AP_{75}$\\
 \hline
 Faster R-CNN & 21.3 &42.0 &19.8  \\
 context/cls & 22.1& 42.9& 20.9\\
 context/loc& 21.4& 41.4& 20.1\\
  \hline
\end{tabular}
\end{center}
\caption{Model performances to demonstrate the impact of global context information on object detection. (context/cls): adding global context information for classification. (context/loc): adding global context information for localization.}
\label{tab2}
\end{table}

Among them, $W_P$ , $W_K$  and  $W_V$  are matrices obtained through learning. As the channels of $V$ and $P$ are corresponding one-by-one, we discarded $W_V$  and observed slightly performance improvement. The distances between vectors represent the relativities of the semantic features. Therefore, the computation is equivalent to calculating the correlation between each part of the region proposal $P$ and each part of the original feature map $G$, and then enhancing the proposal representation with the semantic information of the image according to the degree of correlations. In order to investigate the impact of context information on classification and localization tasks, we separately added the global context information on the two tasks. The results are summarized in Table \ref{tab2}.  As shown, the performance improvement support our suggestions that the performance improvement is mainly reflected on classification.

\subsection{Stride is essential in localization}
\begin{table}[h]\footnotesize
\begin{center}
\begin{tabular}{c|c c c ccc}
 \hline
 & AP& $AP_{50}$& $AP_{75}$ & $AP_{s}$ & $AP_{m}$ & $AP_{l}$\\
 \hline
 detect-F & 21.8 &41.9 &20.7 &5.7 &24.0 &34.3 \\
 -pool/cls & 21.0& 39.8& 20.2 & 6.3 & 23.0 & 31.1\\
 -pool/loc & 22.9& 42.5& 22.7 & 7.1 & 25.9 & 34.6\\
 conv(2,2,2)/loc & 21.4& 41.7& 20.1 & 5.8 & 23.4 & 33.7 \\
 pool(2,2,1)/loc & 23.0& 42.4& 22.9 & 7.2 & 26.3 & 34.6 \\
  \hline
\end{tabular}
\end{center}
\caption{Model performance to demonstrate the impact of global context on detection. (-pool/cls): remove the last pooling layer on classification branch;  (-pool/loc): remove the last pooling layer on localization branch; conv(2,2,2)/loc: replace the last pooling layer  on  localization branch  with a 2*2 convolutional layer with a stride of 2;  pool(2,2,1)/loc: change the stride of the last pooling layer on localization branch from 2 to 1.}
\label{tab3}
\end{table}

Pooling layers are widely used in the convolutional neural networks. Whether in classification, segmentation or detection tasks, the pooling layer serves as an important component of the feature extraction network, which plays an essential role in reducing resolution, decreasing parameters, and transmitting primary information. Nevertheless, the pooling layer brings defects including loss of suboptimal information and inaccurate alignment. ResNet\cite{he2016deep} suggests to use 2*2 convolution kernel with stride of 2 to replace pooling layer for keeping information while reducing the resolution of feature maps. Hinton\cite{sabour2017dynamic} proposes a new design of capsule network to replace the pooling layers. However, currently there is no research quantitatively study the impact of pooling on object detection. We believe that pooling brings different influence on classification and localization tasks. Pooling is important for the classification task, as it enlarges receptive fields and integrates information of the target. On the contrary, the localization task needs smaller strides thus needs fewer pooling layers, as the local deviation caused by the pooling layer will greatly influence the localization accuracy, especially when a high IOU threshold is applied. Therefore, we construct several experiments to prove that the number of pooling layers in current detection networks are more than what they need.

In order to study the impact of pooling on classification and localization tasks separately, we adopted detect-F as the baseline to separate the two tasks since it brings very little parameter growth. The conv5 block of the VGG16 backbone is split for the two tasks. The details are summarized in Table \ref{tab3}. For comparison, the last pooling layers are removed either from the classification branch(-pool/cls) or from localization branch(-pool/loc) in the backbone. To ensure the representation capacity of the network, the convolutional layers following the removed pooling layer are remained.

\begin{figure*}
\begin{center}
\includegraphics[width=0.8\textwidth]{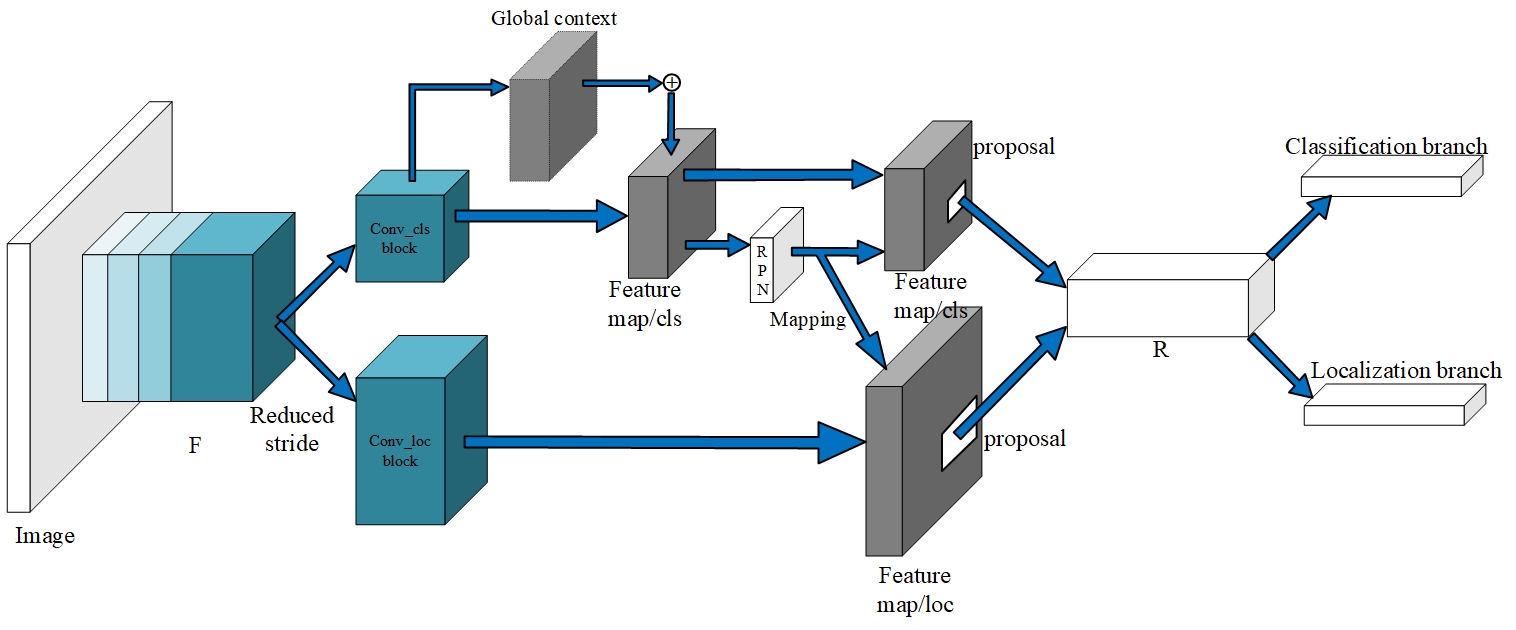}
\end{center}
   \caption{Paradigm of the Gap-optimized Region Based Convolutional Network(G-RCN). (cls) represents classification branch, (loc) represents localization branch. F represents backbone and R represents head.}
\label{fig5}
\end{figure*}

Results show that the pooling layer improves the performance of the classification branch, but on the contrary, it brings performance degradation to the localization branch. Interestingly, we observed that when the pooling layer in the localization branch is removed, not only small objects, but also medium and large objects have performance improvement. On the contrary, there is a significant performance drop when the pooling layer in the classification branch is removed. Therefore, the improvement is not due to decreasing of resolution.

Furthermore, to prove that this performance change is not caused by loss of information during pooling, we adopted a 2*2 convolution kernel with stride of 2 (cov(2,2,2)/loc) and a 2*2 pooling layer with stride of 1 (pool(2,2,1)/loc) to replace the pooling layer in the localization branch as comparison experiments. Interestingly, the 2*2 convolution kernel with stride of 2 has a performance drop even it does not lose information like pooling. On the contrary, a 2*2 pooling layer with stride of 1 significantly improves the performance even it still loses information.

We suggested that this phenomenon is caused by the deviation in localization. Intuitively, the stride determines the distance between two adjacent receptive fields on the feature map. Therefore, the network is not sensible to any changes within the stride, small stride will directly brings localization improvement.  This is reflected in the phenomenon that performance improvement of the localization network after removing the pooling layer is much more significant on $AP_{75}$ than that on $AP_{50}$, and also the removal of the pooling layer can bring the detection performance improvement for all scales of objects. On the contrary, classification prefers larger stride with larger receptive fields to integrate global information.

\section{Gap-optimized Region Based Convolutional Network}

In this section, we combined our discovery in the above section and developed a new paradigm from the general region based object detection models. We achieved this by a new construct that we refer to as Gap-optimized region based convolutional network(G-RCN), the detailed structure is shown in Figure \ref{fig5}. As the stride of the detection model denotes the smallest grain can be recognized, our goal is to limit the stride of detection models for more precise localization. We achieve this without adding any additional modules or information but decreasing the stride in the localization branch in traditional detection models.

\subsection{ResNet-det protocol}
The implementation of ResNet-based detection models originally followed the protocol as in \cite{he2016deep}, in which all layers of conv5 block of ResNet are adopted as the head and attach to the RoI pooling layer. The backbone is constituted by all the layers in the first four blocks and generate feature maps shared for classification and localization. And also, there are experiments \cite{lin2017feature} demonstrating that utilizing the conv5 block in the head can lead to better performance than putting the Conv5 block into the feature extraction network and using shared MLP in the head. Based on Section(cite), we think this protocol is misleading as the localization performance will drop dramatically if put the conv5 layers forward directly. Obviously, placing the Conv5 block on the feature extraction network leads to more powerful and abstract feature representation which is beneficial for classification. The performance drop is mainly caused by the stride on the conv5 block which degrades the localization accuracy, and can be improved easily.

We modified the ResNet-based protocol for detection(named ResNet-det) which simply changes the stride of the first convolutional layer of the Conv5 block from 2 to 1 and moves the block from the head back to the feature extraction network. The head is added back with two fully-connected layers similar with \cite{lin2017feature}. This modification will be proved to bring a significant improvement in the experimental part, though it is a small change.

\subsection{Faster R-CNN}
We also applied G-RCN on the Faster R-CNN with various backbones, we firstly separate the classification and localization tasks from current paradigm since sharing the two tasks is proven to be sub-optimal. To that end, we split these two tasks in the feature extraction network. As multi-task learning is helpful for learning the low-level features, we start splitting at the last few layers of our backbone. On the classification branch, an attention procedure is added to update the feature map according to the correlation between a local part and the global context of the feature map as the global context information is beneficial for classification. On the contrary, since localization requires more detailed information and larger stride usually leads to larger deviation, the stride of the localization branch is reduced(denoted as conv\_loc block). Finally, both classification and localization branches share a common head to prevent a significant increase of parameters.

\section{Dataset and Evaluation}
\begin{table*}[h]\footnotesize
\begin{center}
\begin{tabular}{p{39mm}|p{3mm}p{2mm}p{2mm}p{2mm}p{2mm}p{2mm}p{2mm}p{2mm}p{2mm}p{2mm}p{2mm}p{2mm}p{2mm}
p{2mm}p{2mm}p{2mm}p{2mm}p{2mm}p{2mm}p{2mm}p{2mm}}
 \hline
 Method & \scriptsize{$AP_{70}$}& \scriptsize{aero}& \scriptsize{bike}& \scriptsize{bird}& \scriptsize{boat}& \scriptsize{bottle}& \scriptsize{bus}& \scriptsize{car}& \scriptsize{cat}& \scriptsize{chair}& \scriptsize{cow}& \scriptsize{table}& \scriptsize{dog}& \scriptsize{horse}& \scriptsize{mbike}& \scriptsize{persn}& \scriptsize{plant}& \scriptsize{sheep}& \scriptsize{sofa}& \scriptsize{train}& \scriptsize{tv}\\
 \hline
 VGG16(Baseline) &55.8 &59.0 &60.0 &51.6 &39.0 &43.9 & 75.3& 68.6& 66.0& 36.4& 59.0& 53.9& 61.6& 67.8& 58.1& 54.3& 17.6& 58.8& 59.2& 62.5& 63.8\\
 VGG16 -pool/loc& 56.6& 60.2& 63.2 & 50.0 & 41.2 & 45.3& 69.1& 70.4& 68.5& 39.6& 63.1& 49.5& 60.8& 67.2& 58.8& 54.8& 22.2& 59.2& 57.3& 65.2&67.2\\
 VGG16 -pool/loc + context/cls (G-RCN) & 58.1& 60.9& 60.8 & 51.8 & 41.2 & 43.5& 73.7& 70.4& 66.0& 40.3& 73.4& 53.5& 62.5& 65.6& 62.5& 56.4& 25.1& 63.6& 60.5& 64.2 & 66.5\\
 \hline
 ResNet101(Baseline) & 60.6& 67.6& 67.7& 58.7& 44.3& 46.2& 76.8& 69.1& 77.0& 40.3& 62.7& 51.9& 65.2& 68.7& 66.8& 57.3& 30.5& 57.4& 71.4& 66.9& 65.5\\
 ResNet101(G-RCN) & 63.0&66.0& 67.1& 61.5& 52.2& 49.3& 77.1& 76.1& 74.1& 42.1& 70.3& 53.8& 74.0& 68.7& 63.2& 58.3& 30.3& 66.6& 68.6& 66.8 & 73.1\\
 \hline
 ResNet50(Baseline) & 55.9& 58.3& 58.6& 44.8& 40.3& 40.0& 76.2& 68.5& 70.5& 34.5& 61.8& 54.0& 61.8& 67.5& 57.0& 54.6& 23.4& 58.0& 60.9& 64.0& 63.3\\
 ResNet50(G-RCN) & 57.9& 66.7& 57.4& 53.7& 46.8& 43.9& 76.1& 69.1& 73.9& 33.2& 66.7& 51.3& 62.3& 64.5& 64.3& 56.1& 27.2& 60.7& 57.8 &63.3 &63.1\\
ResNet50-det(Baseline) &59.5& 61.5& 66.1& 56.9& 44.3& 43.5& 75.9& 69.8& 73.1& 38.9& 61.3& 58.8& 63.7& 68.1& 66.4& 56.9& 23.6& 62.7& 66.4& 66.8& 64.7\\
 \hline
\end{tabular}
\end{center}
\caption{Performance of G-RCN on the PASCAL VOC dataset. Baseline:Faster R-CNN framework without separation of classification and localization.(-pool/loc): separate the two tasks and remove the last pooling layer on the localization branch; (context/cls): adding global context information on classification branch. ResNet50-det reduces the stride and moves the Conv5 block back to the feature extraction network. }
\label{tab4}
\end{table*}

\textbf{Datasets and Metrics} We evaluated our proposed paradigm on the COCO 2014 detection dataset\cite{lin2014microsoft} which contains 80 categories and PASCAL VOC 2007+2012 datasets\cite{everingham2010pascal} which contains 20 categories. On the COCO dataset, we trained our image on the training set of 83,000 images and tested the performance on the validation set of 40,000 images. we adopted standard COCO metrics including AP, $AP_{50}$, $AP_{75}$, $AP_{small}$, $AP_{medium}$, $AP_{large}$ to evaluate model performances. Small, medium, large objects are defined as whose size is less than 32*32, more than 32*32 and less than 96*96, more than 96*96 respectively. On the PASCAL VOC dataset, following the protocol in \cite{girshick2015fast}, we used the VOC 2007 trainval dataset of 5,000 images and VOC 2012 trainval dataset of 11,000 images for training, and tested models on VOC 2007 test set of 5,000 images. Since objects on the VOC dataset are relatively large and easy to detect, we used $AP_{70}$  as the evaluation metric to highlight the performance improvement. As is commonly practiced, we adopted pretrained backbone of VGG and ResNet to verify the generalizability of the G-RCN. The results on the PASCAL VOC dataset is shown in Table \ref{tab4} and the results on the COCO dataset is shown in Table \ref{tab5}.

\begin{table*}
\begin{center}
\begin{tabular}{l|c ccccc}
\hline
 Method & AP& $AP_{50}$& $AP_{75}$ & $AP_{small}$ & $AP_{medium}$ & $AP_{large}$\\
 \hline
 VGG16(Baseline) &21.3 &42.0 &19.8 &5.7 &23.2 &33.3 \\
 VGG16 -pool/loc & 22.9& 42.5& 22.7 & 7.1 & 25.9 & 34.5\\
 VGG16 -pool/loc + context/cls (G-RCN) & 23.3& 43.4& 22.9 & 7.8 & 26.5 & 34.6\\
 \hline
 ResNet101(Baseline)  &22.8 &43.7 &21.7 &5.9 &24.6 &37.0 \\
 ResNet101 (G-RCN)& 25.3& 44.8&  25.8&  7.7&  28.6& 39.4\\
\hline
ResNet50(Baseline) & 20.8& 40.9& 19.3& 5.4& 22.7& 33.4\\
ResNet50(G-RCN)  &22.7& 42.0& 22.5&  6.9&  26.1& 35.4 \\
ResNet50-det(Baseline) &22.3& 43.2& 21.0&  5.9&  24.2&  35.6 \\
\hline
\end{tabular}
\end{center}
\caption{Performance of G-RCN on the COCO dataset. Baseline: Faster R-CNN framework without separation tasks.(-pool/loc): separate the two tasks and remove the last pooling layer on localization branch; (context/cls): adding global context information on classification branch. ResNet50-det reduces the stride and moves the Conv5 block back to the feature extraction network.\footnote{New results will be updated soon.}}
\label{tab5}
\end{table*}

\textbf{Implementation Details} We implemented our models and conduct experiments on Tensorflow\cite{abadi2016tensorflow}.  The shorter side of input image is resized to 600 pixels during preprocessing. Each mini-batch contains 1 image, the RPN is trained with batchsize of 256 and Fast R-CNN is trained with 128 Region of interest (ROI)s per batch. SGD is adopted with a momentum of 0.9. We tested our models with NMS of 0.3 and ROIs as 300 per image. For COCO dataset, the learning rate is set as 0.001 for the first 240k iterations and 0.0001 for the rest 80k iterations, totally 320k iterations. Anchors with five sizes and three aspect ratios are adopted. For PASCAL VOC dataset, the learning rate is set as 0.001 for the first 150k iterations and 0.0001 for the rest 30k iterations, totally 180k iterations. Anchors with three sizes and three aspect ratios are adopted.

\subsection{G-RCN  with VGG backbone}
We firstly applied G-RCN on the Faster R-CNN with VGG16 backbone.  Same as that in the above section, the separation starts at the conv5 block with the last pooling layer on the localization branch removed (-pool/loc). An attention procedure is then applied on the classification branch to add the global context information on the feature maps (+context/cls). As shown in Table \ref{tab4} and Table \ref{tab5} , the implementation of G-RCN receives 2.3 $AP_{70}$ improvement on the  PASCAL VOC and 2.0 $AP$ improvement on the COCO dataset. Interestingly, the performance boosting on the COCO dataset is mainly contributed by separation the two tasks and removing the pooling layer, while the performance boosting on the PASCAL VOC dataset is mainly due to adding the global context information. This may because that there are more small objects on the COCO dataset, which are more sensitive to the stride. On the contrary, the objects on the PASCAL VOC dataset are relatively large, therefore the overall performance may more depends on classification.
\subsection{G-RCN with ResNet backbone}
In this section, we adopted the G-RCN on the Faster R-CNN with RestNet50, ResNet101 backbone, and proposed a way to separate these two tasks. Separation for ResNet101 starts at the last 6 bottlenecks in the conv4 block, and the conv5 block remains to be used in the head as original. The first convolutional layer of the first bottleneck in conv4 block, which originally has a stride of 2, is modified to a stride equal to 1 for the localization branch.  But both classification and localization branches share the same kernels of the first 17 bottlenecks of the conv4 block. Similarly, the separation for ResNet50 applies on the last 2 bottlenecks of the Conv4 block and the stride of the first convolutional layer of the Cov4 block is reduced from 2 to 1. Detailed architectures will be provided in the supplementary material.

As shown in Table \ref{tab4} and Table \ref{tab5}, the implementation of G-RCN on ResNet101 receives 2.4 $AP_{70}$ improvement on the  PASCAL VOC dataset  and 2.5 $AP$ improvement on the COCO dataset. The G-RCN also brings  2.0 $AP_{70}$ improvement on the  PASCAL VOC dataset  and 1.9 $AP$ improvement on the COCO dataset for ResNet50.

\subsection{Evaluation of ResNet-det}
\emph{ResNet-det}  follows the modification we proposed above, we performed RoI Align with an output size of $7\times7$ rather than $14\times14$ to save memory footprint on the conv5 layers which modified the stride to 1 for advancing localization performance(these two pooling brings similar performance). Surprisingly, as shown in Table \ref{tab4} and Table \ref{tab5}, ResNet50+ brings 3.6 $AP_{70}$ improvement on PASCAL VOC dataset and 1.5 $AP$ improvement COCO dataset. We also attempt to replace the fc layers with two convolution layers or conv5 block of ResNet but observed no improvement.

\section{Conclusion}
In this work, we presented an analysis to separate the classification and localization tasks for the region based object detection model and study the effect on the two tasks separately. we explored that: (1) Sharing the high-level features for the classification and localization tasks is sub-optimal; (2) Large stride is beneficial for classification but harmful for localization; (3) Global context information could improve the performance of classification. Based on the analysis,  a new paradigm(G-RCN) was proposed to separate the classification and localization tasks and optimizing the gap between them. Experimental results on the PASCAL VOC dataset and COCO dataset demonstrated the significant performance improvement in object detection when applying G-RCN. We further explored that current protocol which places the Conv5 block of ResNet as the head is misleading, one can greatly improve the performance by simply reducing the stride and moving the Conv5 block back to the feature extraction network. In the future, we would like to further improve the ResNet based object detection model by combining this strategy with the implementation of G-RCN, we would also like to extend the G-RCN to apply on the other models such as YOLO, FPN. Furthermore, there may also exists conflict between classification and segmentation tasks, we would also like to explore that.

\bibliographystyle{icml2019}
\bibliography{ref}

\end{document}